\documentclass{article}
\usepackage{arxiv}

\usepackage{graphicx}
\usepackage{algpseudocode}
\usepackage{amsmath}
\usepackage{cite}
\usepackage{float}
\usepackage{amsmath,amssymb,amsfonts}
 
\usepackage{subcaption}
\usepackage{graphicx}
\usepackage{textcomp}
\usepackage{xcolor}
\usepackage{mathtools}
\usepackage{array}
\usepackage{comment}
\usepackage{fullpage}
\usepackage[utf8]{inputenc}
\usepackage[T1]{fontenc}
\usepackage{url}
\usepackage{amsfonts}
\usepackage{physics}
\usepackage{caption}
\usepackage{float}
\usepackage{subcaption}
\usepackage{mathtools}
\usepackage{comment}
\usepackage[title]{appendix}

\usepackage{amsmath,amsfonts,amssymb,amsthm}
\usepackage{nicefrac}
\usepackage{booktabs}

\usepackage{algorithm}
\usepackage{algpseudocode}
\usepackage{geometry}
\usepackage{svg}
\usepackage{amsmath,amsfonts}
\usepackage{amssymb,amsthm}
\usepackage{bm}
\usepackage{xcolor}
\usepackage{multirow}
\usepackage{multicol}
\usepackage{pifont}
\usepackage{booktabs}
\usepackage{appendix}
\usepackage[colorlinks=true, allcolors=blue]{hyperref}
\usepackage{amsthm}

\usepackage{enumitem,amssymb}
\newlist{todolist}{itemize}{2}
\setlist[todolist]{label=$\square$}
\usepackage{pifont}
\newcolumntype{P}[1]{>{\centering\arraybackslash}p{#1}}

\hypersetup{
    colorlinks=true,
    linkcolor=blue,
    filecolor=magenta,      
    urlcolor=cyan,
}

\title{A Particle-based Sparse Gaussian Process Optimizer}

\author{Chandrajit Bajaj \\
Oden Institute\\
University of Texas at Austin \\
 \texttt{bajaj@cs.utexas.edu}
\And Omatharv Bharat Vaidya \\
  Oden Institute \\
  University of Texas at Austin\\
  \texttt{omatharv.vaidya@austin.utexas.edu} \\
   \And Yi Wang \\
   Oden Institute \\
  University of Texas at Austin\\
   \texttt{panzer.wy@utexas.edu}
   }

\begin{document}
\date{}
\maketitle

\begin{abstract}
Task learning in neural networks typically requires finding a globally optimal minimizer to a loss function objective. Conventional designs of swarm based optimization methods apply a fixed update rule, with possibly an adaptive step-size for gradient descent based optimization. While these methods gain huge success in solving different optimization problems, there are some cases where these schemes are either inefficient or suffering from local-minimum. We present a new particle-swarm-based framework utilizing Gaussian Process Regression to learn the underlying dynamical process of descent. The biggest advantage of this approach is greater exploration around the current state before deciding a descent direction. Empirical results show our approach can escape from the local minima compare with the widely-used state-of-the-art optimizers when solving non-convex optimization problems. We also test our approach under high-dimensional parameter space case, namely, image classification task.

\end{abstract}
\section{Introduction}
\label{sec:introduction}

Gradient Descent (GD), a discrete-time iterative scheme, was first introduced in \cite{cauchy1847gd} to solve unconstrained optimization problem. Since then, numerous schemes (using the core principle of gradient descent) have been invented, studied theoretically, and analyzed. For instance, Stochastic Gradient Descent (SGD), proposed by \cite{robbins1951sgd}, quickly became of the most essential optimization algorithms with the rise of machine learning. Some popular methods developed afterwards include momentum method\cite{qian1999momentum}, AdaGrad \cite{duchi2011adaptive}, AdaDelta \cite{zeilar2012adadelta}, RMS-Prop \cite{hinton2012neural}, Adam \cite{kingma2014adam}, Nadam \cite{dozat2016nadam}, AMSGrad \cite{reddi2018convergence}, etc. 

If $\bm\theta$ represents the parameters of the deep network $\mathcal M(\bm\theta)$, the goal of a general task-learning problem is to learn a task $T$ under the unconstrained optimization setting. The task-learning problem can be formulated as optimizing a loss function $\mathcal{L} (\boldsymbol{\theta})$ that best represents the task and finding the best parameter $\bm\theta^{best}$ via the stochastic optimization problem:
\begin{equation} \label{eq: soo}
\bm\theta^{best} = \min_{\bm\theta} \mathcal L(\bm\theta) = \min_{\bm\theta} \mathbb{E}_{\boldsymbol{w}} [\mathcal L(\bm \theta, \boldsymbol{w})],
\end{equation}where, the variable $\boldsymbol{w}$ represents a random variable whose samples can be observed or generated \cite{liu2021stochastic}. This is also called as empirical risk minimization. In this research, as an example, we will consider the case of supervised machine learning for solving a classification problem in images. The model $\mathcal M$ represented by the parameters $\bm\theta$, is a mapping from features to labels. In this case, $\bm{w}$ represents the sampling of features and labels from the data distribution. A simple way to solve optimization problem \eqref{eq: soo} is to use sample-average approximation, i.e. given $N$ i.i.d. samples $\{\boldsymbol{w}^j \}_{j=1}^N$ of the features and labels, the optimization problem is framed as an empirical approximation of \eqref{eq: soo}:
\begin{equation}
\bm\theta^{best} = \min_{\bm\theta}\frac{1}{N} \sum_{i=1}^N \mathcal L(\bm\theta, \boldsymbol{w}^j).
\end{equation}An alternative way is stochastic approximation. SGD \cite{robbins1951sgd} utilizes point-wise stochastic estimates of gradients of the cost function via samples of the data distribution for solving \eqref{eq: soo}:
\begin{equation}
\bm\theta_{t+1} = \bm\theta_t + \alpha_t \nabla_{\bm\theta} \mathcal L(\bm\theta_t, \boldsymbol{w}_t).
\end{equation}where, $\alpha_t$ is the step-size and the gradient $\mathcal L(\bm\theta_t, \boldsymbol{w}_t)$ is evaluated at a single-sample, making it fairly cheap. A variant of SGD is batch gradient descent:
\begin{equation}
\bm\theta_{t+1} = \bm\theta_t - \frac{\alpha_t}{|S_k|} \sum_{j \in S_k} \nabla_{\bm\theta} \mathcal L(\bm\theta_t, \boldsymbol{w}_j)
\label{eq:sgd:vanilla}
\end{equation}where, $S_k$ is a mini-batch sample from $\{1,2,3...,N\}$ of size $|S_k|$, which considers mini-batch samples rather than one sample in SGD \cite{ruder2016overview, hinton2012neural}.

Our goal is to develop an optimization algorithm that betters the current state-of-the-art algorithms for task-learning, and empirically showing these results for the classification task. We note that one of the biggest disadvantages of the widely used methods is that they compute the gradients at a single point, which is usually the state of the network parameters at that moment, and use this information for deciding the descent direction. One of the biggest disadvantages of this approach is that there isn't sufficient exploration around this point, which a few particle-swarm optimization methods have shown. We take an inspiration from particle-swarm optimization to utilize particles for sampling more than one gradients and then use this information in a productive way to model the gradient descent dynamics. A naive approach would be to initialize particles every iteration using a normal distribution around the current parameters' state, sample gradients at these particle locations and to take an average of these gradients as the approximate gradient for the network parameters. While this technique ensures exploration, it's clear that it won't be accurate in case of a tough parameter terrain. Hence, we propose a novel idea of using Gaussian Process Regression \cite{williams1995gaussian} to effectively model the pattern shown by these gradients. This approach ensures approximate gradients and sufficient exploration to show the direction of descent towards the global minima instead of being stuck in a local minima. If the optimizer is stuck in a narrow local minima, there is a chance of having few particles discovering a direction of getting out resulting in a likely overall prediction by the Gaussian Process model to follow that direction. We call our technique ParticleGP. 

We demonstrate the effectiveness of our algorithm by empirical results on widely-known quadratic non-convex optimization problems and on imaging-based classification tasks. The paper is arranged as follows: Section \ref{sec:Relatedwork} discusses related work in particle-based methods and gradient-descent; Section \ref{sec:algorithm} specifies our approach; Section \ref{sec:results} demonstrates the results of comparison of ParticleGP with other state-of-the-art optimizers; Finally, Section \ref{sec:limitations} discusses the limitations and opportunities of improvement.


\section{Related Work}
\label{sec:Relatedwork}

\subsection{GD-based Methods}

GD-based methods adopt a calculation of parameter update $\bm{\delta}_t$ using gradient information at each time step, i.e., if $\bm\theta_{t}$ are the network parameters, they are updated as:
\begin{equation}
\bm\theta_{t} = \bm \theta_{t-1} + \bm\delta_t
\end{equation}Prior techniques in optimization provide a mechanism to compute the best $\bm{\delta}_t$ for fastest convergence to a global minima. GD simply follows the gradient and is scaled by learning rate. Two common tools to improve GD are the sum of gradients (called as the \textit{first moment}) and the sum of the gradients squared (called as the \textit{second moment}). Momentum \cite{qian1999momentum} uses the first moment with a decay rate to gain speed, whereas AdaGrad \cite{duchi2011adaptive} uses the second moment with no decay to deal with sparse features. RMSProp \cite{hinton2012neural} uses the second moment with a decay rate to improve it's rate of convergence over AdaGrad. The fairly popular Adam \cite{kingma2014adam} uses both first and second moments, and is generally regarded as the best choice. Nadam \cite{dozat2016nadam} utilizes Nesterov-acceleration over the Adam scheme.

\subsection{Particle-based Methods}

Compared to GD-based methods, particle-swarm based methods rely on functional evaluations rather than gradient computations for deciding the best direction of descent. Several previous research papers have developed Particle Swarm techniques for optimization purposes. One of the original works was by \cite{kennedy1995particle}, which introduced Particle Swarm Optimization (PSO). Each particle in the swarm has information about: the best location that it has visited, called $\bm\theta_{best}$ and the best location that any particle has visited overall, called the $\hat{\bm\theta}_{best}$. Every particle follows the following dynamical system, wherein it's position and velocity are updated:
\begin{align}
    \bm\delta_{t+1} &= w \bm\delta_t + c_1 r_1 (\bm\theta_{best} - \bm\theta_t) + c_2 r_2 (\hat{\bm\theta}_{best} - \bm\theta_t) \\ 
    \bm\theta_{t+1} &= \bm\theta_t + \bm\delta_{t+1}
\end{align}The position of the particles directly corresponds to the value of parameters i.e. dimension of parameter space is equal to dimension of swarm. Let position of the $i^{th}$ particle at time step $t$ be given by $\bm\theta^i_t$. We have: $\bm \theta^i_{best} =  \arg \min_{t} \mathcal L (\bm \theta^i_{t})$ and $\hat{\bm \theta}_{best} =  \arg \min_{i,t} \mathcal L (\bm \theta^i_{t})$, where $\mathcal L$ is the Lagrangian. The vectors $\mathbf{\delta}$ and $\theta$ represent the velocity and position of the particle respectively, $w$ is the inertia term, $c_1$ and $c_2$ are the relative weights given to cognitive learning and social learning respectively, whereas $r_1$ and $r_2$ are random points drawn from the uniform probability distribution $U(0,1)$ as the damping term. After several iterations, the swarm collectively moves towards the minima as required.
Since then, a lot of papers have focused on swarm intelligence. \cite{xiang2007improved} created a slight modification in the iteration scheme for the velocity by adding a momentum term:
\begin{align}
    \bm\delta_{t+1} &= (1-\lambda)[\bm\delta_t + c_1 r_1 (\bm\theta_{best} - \bm\theta_t) + c_2 r_2 (\hat{\bm\theta}_{best} - \bm\theta_t)] + \lambda \bm\delta_{t-1}
\end{align}where, $\lambda$ denotes the momentum factor. This update helped improve performance by giving a weight to past velocities as well, to relieve excessive oscillation. EM-PSO \cite{mohapatra2021adaswarm} uses an additional momentum term $\mathbf{M}$ to keep track of the exponential average of previous velocities:
\begin{align}
    \bm M_{t+1} &= \beta \bm M_t + (1-\beta) \bm \delta_t \\
    \bm\delta_{t+1} &= \bm M_{t+1} + c_1 r_1 (\bm \theta_{best} - \bm\theta_t) + c_2 r_2 (\hat{\bm \theta}_{best} - \bm\theta_t) \\
    \bm\theta_{t+1} &= \bm\theta_t + \bm\delta_{t+1}
\end{align}This adds flexibility to the task of exploration better than M-PSO and ensures faster convergence. One of the key aspects that the above PSO techniques had lacked was convergence to global optimum. \cite{vaidyahmc} uses a single Hamiltonian Monte Carlo (HMC) particle for effectively searching the optimization space and ensures convergence to global optimum while retaining the benefits of EM-PSO by using $N$ EM-PSO particles. The table \ref{table:1} summarizes different singe-agent prior works; whereas the table \ref{table:2} summarizes different particle-swarm based optimizers for single-task learning.


\begin{table}[h!]
\centering
\begin{tabular}{||c | c ||} 
 \hline
 Name & Scheme to compute $\bm\delta_t$ \\ [0.5ex] 
 \hline\hline
 GD \cite{cauchy1847gd}& $\begin{aligned} \bm\delta_t &= - \alpha_t \cdot \nabla_{\bm\theta} \mathcal L(\bm\theta_{t-1}) \end{aligned}$  \\
 \hline
 Momentum \cite{qian1999momentum} & $\bm\delta_t = - r_t \cdot \nabla_{\bm\theta} \mathcal L(\bm\theta_{t-1}) + \bm\delta_{t-1} \cdot \beta_1$ \\
\hline
 AdaGrad \cite{duchi2011adaptive} & $\begin{aligned}
  \boldsymbol{v}_{t-1} &= \nabla_{\bm\theta} \mathcal{L}(\bm\theta_{t-1})^2+ \boldsymbol{v}_{t-2}\\
 \bm\delta_t &= -r_t \cdot \frac{\nabla_{\bm\theta} \mathcal{L}(\bm\theta_{t-1})}{\sqrt{\boldsymbol{v}_{t-1}}}
 \end{aligned}$\\
\hline
 RMSProp \cite{hinton2012neural} & $\begin{aligned} \boldsymbol{v}_{t-1}  &= \nabla_{\bm\theta} \mathcal L(\bm\theta_{t-1})^2 \cdot (1- \beta_2) + \boldsymbol{v}_{t-2} \cdot \beta_2 \\
  \bm \delta_t &= -r_t \cdot \frac{\nabla_{\bm\theta} \mathcal L(\bm\theta_{t-1})}{\sqrt{\boldsymbol{v}_{t-1}}}\end{aligned}$ \\
  \hline
 Adam \cite{kingma2014adam} & $\begin{aligned}
  \boldsymbol{m}_{t-1} &= \nabla_{\bm \theta} \mathcal L(\bm\theta_{t-1}) \cdot (1-\beta_1) + \boldsymbol{m}_{t-2} \cdot \beta_1\\ 
  \boldsymbol{v}_{t-1} &= \nabla_{\bm\theta} \mathcal L(\bm\theta_{t-1})^2 \cdot (1- \beta_2) + \boldsymbol{v}_{t-2} \cdot \beta_2  \\
  \hat{\boldsymbol{m}}_{t-1} &=  \frac{\boldsymbol{m}_{t-1}}{1 - \beta_1}\\
  \hat{\boldsymbol{v}}_{t-1} &= \frac{\boldsymbol{v}_{t-1}}{1 - \beta_2} \\
  \bm\delta_t ~&=~ -r_t \cdot \frac{\hat{\boldsymbol{m}}_{t-1}}{ \sqrt{\hat{\boldsymbol{v}}_{t-1}}}
 \end{aligned}$\\
 \hline
 Nadam \cite{dozat2016nadam} & $\begin{aligned}
  \boldsymbol{m}_{t-1} &= \nabla_{\bm\theta} \mathcal L(\bm\theta_{t-1}) \cdot (1-\mu_{t-1}) + \boldsymbol{m}_{t-2} \cdot \mu_{t-1}\\ 
  \boldsymbol{v}_{t-1} &= \nabla_{\bm\theta} \mathcal L(\bm\theta_{t-1})^2 \cdot (1- \nu) + \boldsymbol{v}_{t-2} \cdot \nu\\
  \hat{\boldsymbol{m}}_{t-1} &=  \mu_{t} \cdot \frac{\boldsymbol{m}_{t-1}}{1 - \prod_{i=1}^{t} \mu_i} + (1-\mu_{t}
 ) \cdot \frac{\nabla_{\bm\theta} \mathcal L(\bm\theta_{t-1})}{1 - \prod_{i=1}^{t} \mu_i}\\
 \hat{\boldsymbol{v}}_{t-1} &= \nu \cdot \frac{\boldsymbol{v}_{t-1}}{1 - \nu^{t-1}}\\
 \boldsymbol{\delta}_t ~&=~ -r_t \cdot \frac{\hat{\boldsymbol{m}}_{t-1}}{ \sqrt{\hat{\boldsymbol{v}}_{t-1}}}
 \end{aligned}$\\
 \hline
\hline
\end{tabular}%
\caption{List of single-agent prior optimizers}
\label{table:1}
\end{table}

\begin{table}[h!]
\centering
\begin{tabular}{||c | c ||} 
 \hline
 Name & Scheme to compute $\delta_t$ \\ [0.5ex] 
 \hline\hline
 PSO \cite{kennedy1995particle} & $\begin{aligned}\bm\delta_t &= w\bm\delta_{t-1} + c_1 r_1 (\bm\theta_{best} - \bm\theta_{t-1}) + c_2 r_2 (\hat{\bm \theta}_{best} - \bm\theta_{t-1}) \end{aligned}$\\
 \hline
 M-PSO \cite{xiang2007improved} & $\begin{aligned} \bm \delta_t &= (1-\lambda)[\bm\delta_{t-1} + c_1 r_1 (\bm\theta_{best} - \bm\theta_{t-1})  + c_2 r_2 (\hat{\bm\theta}_{best} - \bm\theta_{t-1})] + \lambda \bm\delta_{t-1} \end{aligned}$ \\
\hline
 EM-PSO \cite{mohapatra2021adaswarm} & $\begin{aligned}
 \boldsymbol{M}_{t} &= \beta \boldsymbol{M}_{t-1} + (1-\beta) \bm\delta_{t-1}\\
 \bm\delta_t &= \boldsymbol{M}_{t} + c_1 r_1 (\bm\theta_{best} - \bm\theta_{t-1}) + + c_2 r_2 (\hat{\bm\theta}_{best} - \bm\theta_{t-1})
 \end{aligned}$\\
\hline
 REM-PSO \cite{mohapatra2021adaswarm} & $\begin{aligned}
 \boldsymbol{M}_{t} &= \beta \boldsymbol{M}_{t-1} + (1-\beta) \bm\delta_{t-1}\\
 \bm\phi_1 &= \text{diag} (c_{1,1}r_{1,1},  c_{1,2}r_{1,2}, c_{1,3}r_{1,3} , \cdots , c_{1,d}r_{1,d})\\
 \bm\phi_2 &= \text{diag} (c_{2,1}r_{2,1},  c_{2,2}r_{2,2}, c_{2,3}r_{2,3}, \cdots , c_{2,d}r_{2,d})\\
 \bm\delta_t &= \boldsymbol{M}_{t} + \boldsymbol{A}^T \bm\phi_1 \boldsymbol{A} (\bm\theta_{best} -\bm \theta_{t-1}) + \boldsymbol{A}^T \boldsymbol{\phi_2} \boldsymbol{A} (\hat{\boldsymbol{\theta}}_{best} - \bm\theta_{t-1})
 \end{aligned}$ \\
 \hline
 HMC-PSO \cite{vaidyahmc} & uses $N$ EM particles that follow EM-PSO iteration scheme;\\
& and $1$ particle that does HMC sampling for exploration of state space\\
\hline
\hline
\end{tabular}
\caption{List of particle-swarm based optimizers}
\label{table:2}
\end{table}


We note that optimizer methods in table \ref{table:1} evaluate the direction of descent by doing gradient evaluations of the cost function $\mathcal L$ at different points. Meanwhile, for the optimizers in table \ref{table:2}, the best direction to explore state space and exploit towards the minima is captured stochastically via an estimate of the distribution of functional evaluations. For instance, the terms $(\bm\theta_{best} - \bm\theta_{t-1})$ and ($\hat{\bm\theta}_{best} - \bm\theta_{t-1}$) in PSO-based optimizers act as an equivalent to gradients in standard optimizers since they estimate the direction of convergence. Whereas, the exponential momentum in EM-PSO stores and utilizes prior-descent directions similar to Adam and RMS-Prop's iteration schemes. Thus, what a single-agent does in standard optimizers is built intrinsically in swarm intelligence.

While it's clear that there is a deeper connection between swarm-based methods and single-agent optimizers, there is a crucial aspect lacking in both. As discussed in section \ref{sec:introduction}, there isn't an exploration component in single-agent optimizers which might uplift their state from a local minima, whereas for particle-based methods, an essential part that is missing in their design is their ability to compute and utilize gradients. Only using functional evaluations is sub-optimal since we are aware that gradients provide the direction of the fastest decrease of the objective functional at the given state. We bridge gaps in both these techniques by designing a particle-based gradient descent scheme using Gaussian Process Regression. It is able to compute gradients in its surrounding and model the best direction of descent by making sense of all the useful information. We go into the mathematical details in the next section.


\section{Our contribution: ParticleGP}
\label{sec:algorithm}

We are motivated from searching the parameter space under PSO framework and propose a particle-based optimization algorithm using multi-output Gaussian process (called \textit{ParticleGP}) for simulating the dynamical system of parameter updates. As discussed in section \ref{sec:introduction}, one of the primary advantages of using such a technique is that the optimizer can process essential information about the geometry of the parameter space from its local neighbourhood and make an accurate prediction on the objectively best direction to take a descent step. The current state-of-the-art optimizers utilize direct or stochastic gradients at the given point, which while it helps taking the locally best decision, does not paint the entire picture. Our proposed ParticleGP is the most effective method in optimization problems with several local minima and non-convex behaviour: a scenario in which most state-of-the-art optimizers fail to reach the global optima.
 The setup for our problem is as follows: we consider a particle swarm $\{ \bm\theta_t^i \}_{i=1}^{N}$ with swarm size $N$ and an agent $\bm\theta_t^0$; where $i$ represents the specific particle and $t$ is the time-step index. These particles are sampled from a normal distribution, $\bm\theta_{t+1}^i \sim \mathcal N (\bm\theta^0_t, \bm\Sigma_t)$. The agent follows a gradient-descent based dynamical system whereas the surrounding particles assist the agent by proving crucial approximated information about the underlying geometry of the space. Let $h(\bm\theta_t)$ define a random variable depicting a stochastic process, which is Gaussian in nature. It represents the fit of the dynamical process of gradient descent by predicting the gradients. After every iteration $t$, a multi-output Gaussian process model uses information about the optimization manifold from the current particles' positions $\{\bm\theta_{t}^i\}$ and the gradients at these locations $\{\nabla_{\bm \theta} \mathcal L(\bm\theta_{t}^i) \}$ as training data to fit the change $\bm{\delta}_t$ for the main agent. We know that $\bm \theta_t \in \mathbb{R}^{d}$. Let $\bm\Theta_t$ denote the collection of location of all particles as: $\bm\Theta_t = [\bm\theta_{t}^1 \ \bm\theta_{t}^2 \ ...  \ \bm\theta_{t}^N]^T$, and the location of the agent is the testing data: $\bm\theta^0_{t} = [\theta^{0}_{1,t} \ \theta^{0}_{2,t} \ ...\ \theta^{0}_{d,t}]$. Hence, $\bm\Theta_t \in \mathbb{R}^{N \text{ x } d}$ and $\bm \theta_t^0 \in \mathbb{R}^{d}$. The complete training data collection is given by: $(\boldsymbol{X}_t, \boldsymbol{Y}_t) = (\bm\Theta_t, \nabla_{\bm \theta} \mathcal L(\bm\Theta_t) )$, whereas the point on which the model would be tested to calculate approximated gradient is:  $\bm\theta^0_{t}$. 
 
 For any Gaussian Process Regression problem, we require a prior and the likelihood function. Since the optimization is over a $d$-dimensional space, the multi-output Gaussian process model utilizes $d$ different Gaussian process models for representing and prediction information at each dimension \cite{liu2018remarks}. The random variable $h(\bm X_t)$ can hence be represented as: $h(\bm X_t) = [h_1(\bm X_{1,t}), \ h_2(\bm X_{2,t}), \ ... , \  h_d(\bm X_{d,t})]$, where $\bm X_{i,t}$ represents location of all particles in the $i^{th}$ dimension at the $t^{th}$ iteration. Hence, to summarize, the relationship between observations, i.e. the gradients ($\boldsymbol{Y}_t$), and the output of the Gaussian Process Regression model ($h_{i} (\boldsymbol{X_{i,t}})$) is:
 \begin{equation}
 \bm Y_{i,t}(\bm X_{i,t}) = \boldsymbol{I}_N h_i(\bm X_{i,t}) + \boldsymbol{\epsilon}_i, \ \ \text{where} \ \boldsymbol{\epsilon}_i \sim \mathcal{N} (\boldsymbol{0}, \eta^2 \boldsymbol{I}_N); \ \ \forall \ i \in \{1,2,3..., d.\}
 \end{equation}The marginalized Gaussian process prior distribution is given by:
\begin{equation} \label{eq:prior}
\begin{bmatrix}
h_i(\bm X_{i,t}) \\
h_i(\theta^{0}_{i,t})
\end{bmatrix} \sim \mathcal{N} (\begin{bmatrix}
\boldsymbol{M}_{i,t}\\
m_{i,t}
\end{bmatrix}
, \begin{bmatrix}
k(\bm X_{i,t}, \bm X_{i,t}) & k(\bm X_{i,t}, \theta^{0}_{i,t})\\
k(\theta^{0}_{i,t}, \bm X_{i,t}) & k(\theta^{0}_{i,t}, \theta^{0}_{i,t})
\end{bmatrix}), \ \ \forall \ i \in \{1,2,3..., d\}
\end{equation}where, $\boldsymbol{M}_{i,t} =  [\boldsymbol{m}_{i,t}^0 \ \boldsymbol{m}_{i,t}^1 \ \boldsymbol{m}_{i,t}^2 \ ...  \ \boldsymbol{m}_{i,t}^N]^T$ represents the generalized mean function for predicting gradients for each of the particles on the $i^{th}$ dimension, $m_{i,t}$, similarly, is the mean function for the main agent, $k$ is the kernel function defined for two points $x, x' \in \mathbb{R}$ as:
\begin{equation}
k(x,x') = \sigma^2 \exp (-\frac{(x-x')}{2l^2})
\end{equation}whereas for the vector $\boldsymbol{x} = (x_1, x_2, x_3, ..., x_d) \in \mathbb{R}^d$:
\begin{equation}
k(\boldsymbol{x}, \boldsymbol{x}) = \begin{bmatrix}
k(x_1, x_1) & k(x_1, x_2) & ... & k(x_1, x_d)\\
k(x_2, x_1) & k(x_2, x_2) & ... & k(x_2, x_d)\\
... & ... & ... & ...\\
k(x_d, x_1) & k(x_d, x_2) & ... & k(x_d, x_d)
\end{bmatrix} \ \in \mathbb{R}^{d \ \text{x} \ d}
\end{equation}According to the marginalization rule, we can establish the prior distribution as a multi-variate Gaussian distribution:
\begin{align}
h_i(\bm X_{i,t}) \sim \mathcal{N} (\boldsymbol{M}_{i,t}, k(\bm X_{i,t}, \bm X_{i,t}))\\
h_i(\theta^{0}_{i,t}) \sim \mathcal{N} (m_{i,t}, k(\theta^{0}_{i,t}, \theta^{0}_{i,t})); \ \ \forall \ i \in \{1,2,3..., d\}
\end{align}\cite{consonni2018prior} explores more about the importance of choosing an informative prior, and describes the techniques and assumptions for doing so. Let likelihood at training data points be given by: $p_i(\boldsymbol{X}_{i,t})$ and at test data points be given by: $p_i(\theta^0_{i,t})$. We can now define the posterior distribution for modelling the dynamics using the Bayes rule as:
\begin{equation} \label{eq:bayes}
p_i(h_i(\bm X_{i,t}) | \bm Y_{i,t}) = \frac{p_i( \bm Y_{i,t} | h_i(\bm X_{i,t}))}{\int  p_i( \bm Y_{i,t}) | h_i(\bm X_{i,t})) \ p_i(h_i(\bm X_{i,t})) \ d h_i(\bm X_{i,t})} \ \ \forall \ i \in \{1,2,3..., d\}
\end{equation}We note that the output (in the form of a random variable) $\bm Y_{i,t}(\bm X_{i,t})$ can be represented as a linear transformation of $h_i(\bm X_{i,t})$, i.e.:
\begin{equation}
\bm Y_{i,t}(\bm X_{i,t}) = \boldsymbol{I}_N h_i(\bm X_{i,t}) + \boldsymbol{\epsilon}_i, \ \ \text{where} \ \boldsymbol{\epsilon}_i \sim \mathcal{N} (\boldsymbol{0}, \eta^2 \boldsymbol{I}_N); \ \ \forall \ i \in \{1,2,3..., d\}
\end{equation}Using linear transformation property for Normal distributions, the marginal distribution for the output is:
\begin{equation}
\bm Y_{i,t}(\bm X_{i,t}) \sim \mathcal{N} (\boldsymbol{M}_{i,t}, \boldsymbol{K} + \eta^2 \boldsymbol{I}_N) \ \ \forall \ i \in \{1,2,3..., d\}
\end{equation}where, $\boldsymbol{K} = \begin{bmatrix}
k(\bm X_{i,t}, \bm X_{i,t}) & k(\bm X_{i,t}, \theta^{0}_{i,t})\\
k(\theta^{0}_{i,t}, \bm X_{i,t}) & k(\theta^{0}_{i,t}, \theta^{0}_{i,t})
\end{bmatrix}$. Using this information, Bayes rule can be applied to compute the joint distributions:
\begin{equation}
\begin{bmatrix}
h_i(\theta^{0}_{i,t})\\
\bm Y_{i,t}(\bm X_{i,t})
\end{bmatrix} \sim \mathcal{N} (\begin{bmatrix}
m_{i,t} \\
\boldsymbol{M}_{i,t}
\end{bmatrix}
, \begin{bmatrix}
k(\theta^{0}_{i,t}, \theta^{0}_{i,t}) & k(\theta^{0}_{i,t}, \bm X_{i,t}) \\
k(\bm X_{i,t}, \theta^{0}_{i,t}) & k(\bm X_{i,t}, \bm X_{i,t})+ \eta^2 \boldsymbol{I}_N
\end{bmatrix}), \ \ \forall \ i \in \{1,2,3..., d\}
\end{equation}The posterior distribution is $h_i(\theta^{0}_{i,t}) \ | \ \bm Y_{i,t}(\bm X_{i,t}) \sim \mathcal{N} (\bm \mu_{i,t}^*, \bm \sigma_{i,t}^*)$, where:
\begin{align} \label{eq:updatednormal}
\bm \mu_{i,t}^* = m_{i,t} + k(\theta^{0}_{i,t}, \bm X_{i,t})  (k(\bm X_{i,t}, \bm X_{i,t})+ \eta^2 \boldsymbol{I}_N)^{-1} (\bm Y_{i,t}(\bm X_{i,t}) - \boldsymbol{M}_{i,t})\\
\bm \sigma_{i,t}^* = k(\theta^{0}_{i,t}, \theta^{0}_{i,t}) - k(\theta^{0}_{i,t}, \bm X_{i,t}) (k(\bm X_{i,t}, \bm X_{i,t})+ \eta^2 \boldsymbol{I}_N)^{-1} k(\theta^{0}_{i,t}, \bm X_{i,t})^T
\end{align}While utilizing the Gaussian Regression process, we have the following model parameters:
\begin{itemize}
\item length-scale $l$, from the kernel function, is a single scalar.
\item signal variance $\sigma^2$, also from the kernel function, is a single scalar.
\end{itemize}
It is crucial to note the importance of the role of these model parameters for effectively learning the dynamics. Sub-par values of these hyper-parameters can lead to completely incorrect dynamics (i.e. over-fitting and under-fitting the descent process). Fortunately, we are able to optimize them by solving a separate optimization problem i.e. the log of the marginal likelihood :

\begin{align} \label{eq:logmarginallikelihoodoptim}
\log(p(\bm Y_{i,t}(\bm X_{i,t}))) = \log(\frac{1}{(2 \pi)^{N/2} \text{det} 
(\boldsymbol{K} + \eta^2 \boldsymbol{I}_N)^{1/2}} \exp (- \frac{1}{2} (\bm Y_{i,t}(\bm X_{i,t}) - \boldsymbol{M}_{i,t})^T ( \boldsymbol{K}+ \eta^2 \boldsymbol{I}_N)^{-1} (\bm Y_{i,t}(\bm X_{i,t}) - \boldsymbol{M}_{i,t})))\\
= - \frac{1}{2} \log ( \text{det} 
(\boldsymbol{K} + \eta^2 \boldsymbol{I}_N)^{1/2})  - \frac{1}{2} (\bm Y_{i,t}(\bm X_{i,t}) - \boldsymbol{M}_{i,t})^T ( \boldsymbol{K}+ \eta^2 \boldsymbol{I}_N)^{-1} (\bm Y_{i,t}(\bm X_{i,t}) - \boldsymbol{M}_{i,t}) - \frac{N}{2} \log (2 \pi)
\end{align}
Maximizing the above objective function would help in obtaining the dynamical process that best fits the parameters' behaviour. The approximate gradients for the main agent are then sampled from the updated posterior distribution and coupled together from each of the $d$ Gaussian-processes, i.e. $\hat{\nabla}_{\bm \theta} \mathcal L(\theta_{i,t}^0) \sim \mathcal{N} (\bm \mu_{i,t}^*, \bm \sigma_{i,t}^*)$ and $\hat{\nabla}_{\bm \theta} \mathcal L(\bm \theta_{t}^0) = (\hat{\nabla}_{\bm \theta} \mathcal L(\theta_{1,t}^0), \ \hat{\nabla}_{\bm \theta} \mathcal L(\theta_{2,t}^0), \ ..., \ \hat{\nabla}_{\bm \theta} \mathcal L(\theta_{d,t}^0))$. Finally, the update to the main-agent's/network's parameters is carried out as follows:
\begin{align}
\bm\delta^0_t &= - \alpha_t \cdot \hat{\nabla}_{\bm \theta} \mathcal L(\bm \theta_{t}^0)\\
\bm\theta^0_{t+1} &= \bm \theta^0_{t} + \bm\delta^0_t
\end{align}This procedure can be completed for each iteration until the parameters converge to the optimal value. The algorithm \ref{alg:ParticleGP} summarizes our approach.

\begin{algorithm}[h]
\caption{Particle-based Gaussian-process Regression for gradient descent} 
\label{alg:ParticleGP}
\begin{algorithmic}
\State \textbf{Initialize}:  (i) particle swarm $\{ \bm\theta^{i}_0 \}_{i=1}^N \in \mathbb{R}^d$, (ii) main agent representing weights of the neural network $\bm\theta_0^0 \in \mathbb{R}^d$, (iii) step-size for dynamical system update $\eta_t$, (iv) co-variance matrix for sampling particles $\bm \Sigma_t$, (v) model parameters for multi-output Gaussian process $l$ and $\sigma^2$.
\For{$t=1, 2, 3 ... T$}
		\State{Select $N_t$ particles from the multi-variate normal distribution: $\{\bm\theta_{t+1}^i \}_{i=1}^{N_t} \sim \mathcal N (\bm\theta^0_t, \bm\Sigma_t)$ 
		}
        \State{Obtain gradients at particle locations: $\{\nabla_{\bm \theta} \mathcal L(\bm\theta_{t}^i) \}$}
        \For{$i=1, 2, 3, ..., d$}
        		\State{Initialize the $i^{th}$ Gaussian-process Regression prior distribution using equation 
        		}
                \State{Get the posterior distribution via updated values of $\bm \mu_{i,t}^*$ and $\bm \sigma_{i,t}^*$ computed using equation \eqref{eq:updatednormal}}
                \State{Optimize the log-marginal likelihood function in equation \eqref{eq:logmarginallikelihoodoptim} to get the best model parameters $l$ and $\sigma^2$}
                \State{Get the updated posterior for final values of $\bm \mu_{i,t}^*$ and $\bm \sigma_{i,t}^*$ computed using equation \eqref{eq:updatednormal}}
                \State{Sample from posterior: $\hat{\nabla}_{\bm \theta} \mathcal L(\theta_{i,t}^0) \sim \mathcal{N} (\bm \mu_{i,t}^*, \bm \sigma_{i,t}^*)$}
        \EndFor
        \State{Concatenate values to get approximate gradients: $\hat{\nabla}_{\bm \theta} \mathcal L(\bm \theta_{t}^0) = (\hat{\nabla}_{\bm \theta} \mathcal L(\theta_{1,t}^0), \ \hat{\nabla}_{\bm \theta} \mathcal L(\theta_{2,t}^0), \ ..., \ \hat{\nabla}_{\bm \theta} \mathcal L(\theta_{d,t}^0))$}
        \State{Use GD/Momentum/Adagrad/RMS-Prop/Adam/NAdam for the dynamical system according to the table \ref{table:2} as required. For a simple gradients update (i.e. GD), take: $\bm\delta^0_t = - \eta_t \cdot \hat{\nabla}_{\bm \theta} \mathcal L(\bm \theta_{t}^0)$}
        \State{Update the parameters of the main-agent network using: $\bm\theta^0_{t+1} = \bm \theta^0_{t} + \bm\delta^0_t$}
\EndFor
\end{algorithmic}
\end{algorithm}

\section{Results}
\label{sec:results}

\subsection{Non-Convex Quadratic optimization problems}

\begin{figure}[!tbp]
  \centering
  \subfloat[Levy]{\includegraphics[width=0.5\textwidth]{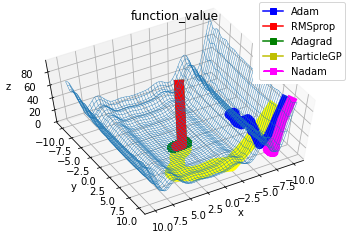}}
  \hfill
  \subfloat[Easom]{\includegraphics[width=0.5\textwidth]{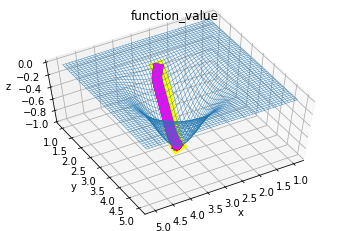}}
  \hfill
  \subfloat[Beale]{\includegraphics[width=0.5\textwidth]{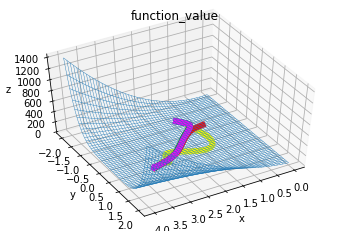}}
  \hfill
   \subfloat[Matyas]{\includegraphics[width=0.5\textwidth]{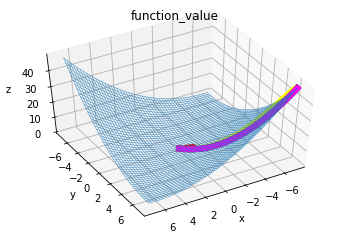}}
  \caption{Sample Quadratic Optimization function mesh-plots: We present the topographic map of the function's space along with the individual trajectory of progress of agents trained on each of the optimizers as they are trained from first epoch. The trajectory is shown in terms of color-coded points representing the model's prediction of the global minima of the function.}
  \label{fig:mesh}
\end{figure}

We tested our algorithm, Particle-GP, on several difficult non-convex problems in $\mathbb{R}^2$. These problems, due to their incredibly complex optimization manifold, were expected to prevent  conventional optimizers converge to the global minima. We considered a set-up in which swarm size is $N = 100$ particles. Hence, the total number of gradient computations for one step of Particle-GP is 100 times  larger than one step in SGD. We sampled particles from a normal distribution with an epsilon ball of radius $\min(0.1, \eta_t)$ centered at the current parameters' state. We ran ParticleGP for $200$ iterations, whereas the rest of the state-of-the-art optimizers for $20000$ iterations on a logistic regression model with parameters as the position of the points in the optimization space. This was done to ensure the \textbf{total number of gradient computations are equal}. The time taken to run state-of-the-art optimizers is \textbf{155} seconds, whereas Particle-GP takes $\textbf{33}$ seconds for the given number of iterations. The table \ref{table:learning_rate} displays the step size ($\eta_t$) setup for the optimizers for each of the experiments. Each optimizer has it's own mechanism to update gradients and due to tough terrains shown by these functions, it becomes necessary to run these optimizers on the settings that favour them the most. Additionally, step-LR is required to ensure that the optimizers don't escape the global minima. Table \ref{table:3} showcases the numerical results of convergence of the model trained using different optimizers. The functions on which experiments were conducted are described in detail on \cite{simulationlib}. Figure \ref{fig:mesh} present a few plots of trajectories of optimizers over different functions. For particleGP, only agent 0 is displayed.

\begin{table}
\centering
\caption{The step-size ($\eta_t$) for experiments on non-convex optimization problems (Experiments for \textbf{bold} functions required a StepLR action with weight decay rate 0.0001 and step size equal to 1 for a single step to shift the learning rate to normal. This was done after the optimizer reached a loss of lower than $10$.)}
\begin{tabular}{||c | c | c | c | c | c ||}
\hline
 Functions \cite{simulationlib} & ParticleGP & Adam & RMSProp & AdaGrad & NAdam \\ [0.5ex] 
 \hline
 Himmelblau & 0.0005 & 0.0005 & 0.0005 & 0.1 & 0.0005  \\
 \hline
 \textbf{Ackley} & 0.095 & 0.5 & 0.5 & 10 & 1\\
 \hline
 Beale & 0.01 & 0.001 & 0.5 & 0.001 & 0.001\\
 \hline
 Goldstein Price & 0.00001 & 0.001 & 0.001 & 0.1 & 0.001 \\
 \hline
 Three Hump Camel & 0.25 & 2.5 & 0.1 & 1 & 2.5 \\
 \hline
 Easom & 0.003 & 0.001  & 0.001  & 0.01 & 0.001\\
 \hline
 Bukin & 0.001 & 0.001 & 0.001 & 0.01 & 0.0001\\
 \hline
  Matyas  & 0.01 & 0.001 & 0.001 & 1 & 0.001 \\
 \hline
  \textbf{Dropwave} & 1 & 1 & 0.25 & 0.1 & 0.01 \\
 \hline
 \textbf{Levy} & 0.08 & 1 & 1 & 10 & 0.5 \\
 \hline
\end{tabular}
\label{table:learning_rate}
\end{table}

\begin{table}
\centering
\caption{Numerical results for non-convex optimization problems (\textbf{bold} indicates that the algorithm converges to local minima instead of global or didn't show signs of converging to the global minima): These numbers represent the euclidean distance between the actual global minimum of the function and what the model trained using these optimizers predicted after the end of it's training. (* -> none of the optimizers converged when initialized on the plane).}
\begin{tabular}{||c | c | c | c | c | c ||}
\hline
\multicolumn{6}{||c||}{L-2 Norm distance from the global optimum} \\
 \hline \hline
 Functions \cite{simulationlib} & ParticleGP & Adam & RMSProp & AdaGrad & NAdam \\ [0.5ex] 
 \hline
 Himmelblau &  0.0046 & 0 & 0 & 0.0003 & 0 \\
 \hline
 Ackley & 1.3968 & \textbf{22.4857} & 1.1985 & \textbf{23.3218} & 0.1148\\
 \hline
 Beale & 0.1194  & 0 & 0.0007 & 0.0001  & 0  \\
 \hline
 Goldstein Price & 0.0362  & 0 & 0 & 0  & 0  \\
 \hline
 Three Hump Camel & 0.0681 & 0.0004 & 0.0707 & 0 & 0.1138\\
 \hline
 Easom* & 0.0028 & 0 & 0.0017 & 0.0002 & 0.0001 \\
 \hline
 Bukin & 0.6403  & \textbf{2.5012} & \textbf{2.5712} & \textbf{2.5033} & \textbf{2.2515}\\
 \hline
  Matyas  & 0.0035 & 0  &  0 & 0.5101 & 0 \\
 \hline
  Dropwave & 1.5624 & 2.051 & 1.9728 & \textbf{7.8506} & \textbf{7.3193} \\
 \hline
 Levy & 0.6005 & \textbf{9.2909} & 1.3523 & 0 & \textbf{12.3008} \\
 \hline
\end{tabular}
\label{table:3}
\end{table}

We observe that for almost all optimization functions, ParticleGP was able to converge to the global minimum. It did better than other state-of-the-art optimizers in terms of the number of functions it found the global optima since the number of times it was successful in being close to the global optimum and not being stuck in a local minimum was one of the highest out of all optimizers, but poorly in terms of finding the exact point. This can be inferred due to ParticleGP using approximated gradients and not the exact ones. The biggest advantage of utilizing ParticleGP is that it uses useful information from particles in the neighbourhood, to find the direction of descent to the global minima; and hence is mostly successful in solving problems with a complex manifold.


\subsection{Classification task}
\begin{figure}[!tbp]
  \centering
  \subfloat[Running Loss]{\includegraphics[width=0.5\textwidth]{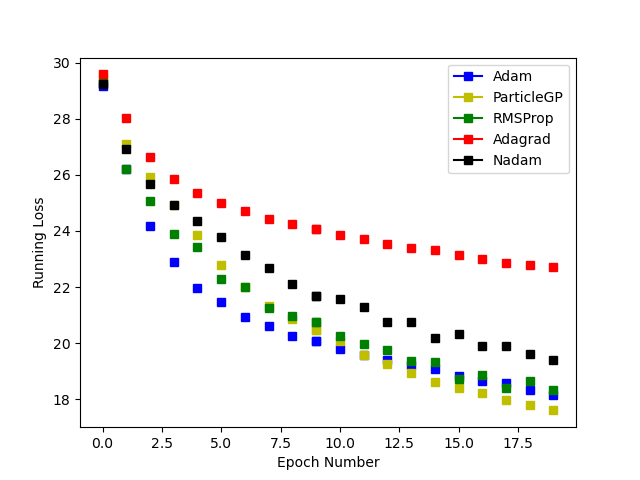}}
  \hfill
  \subfloat[Training Accuracy]{\includegraphics[width=0.5\textwidth]{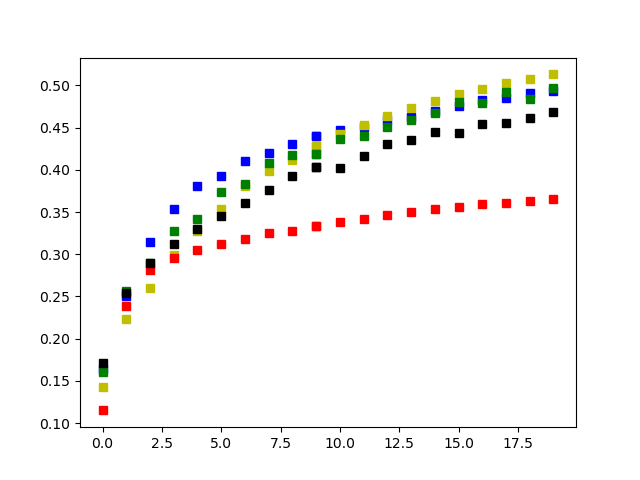}}
  \caption{Performance on the MNIST dataset}
  \label{fig:MNIST}
\end{figure}

\begin{figure}[!tbp]
  \centering
  \subfloat[Running Loss]{\includegraphics[width=0.5\textwidth]{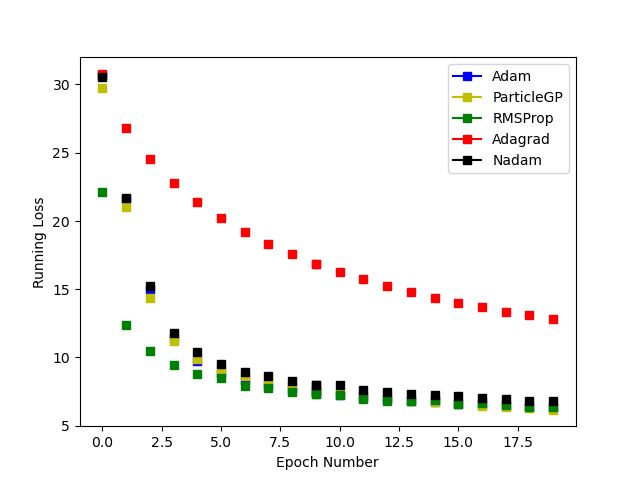}}
  \hfill
  \subfloat[Training Accuracy]{\includegraphics[width=0.5\textwidth]{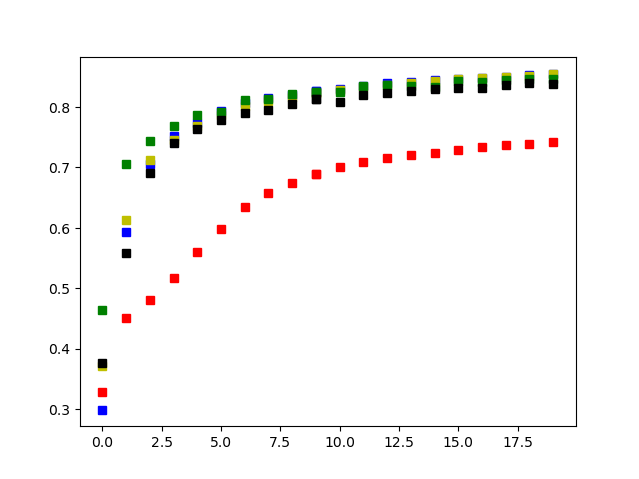}}
  \caption{Performance on the Fashion MNIST dataset}
  \label{fig:FashionMNIST}
\end{figure}

\begin{figure}[!tbp]
  \centering
  \subfloat[Running Loss]{\includegraphics[width=0.5\textwidth]{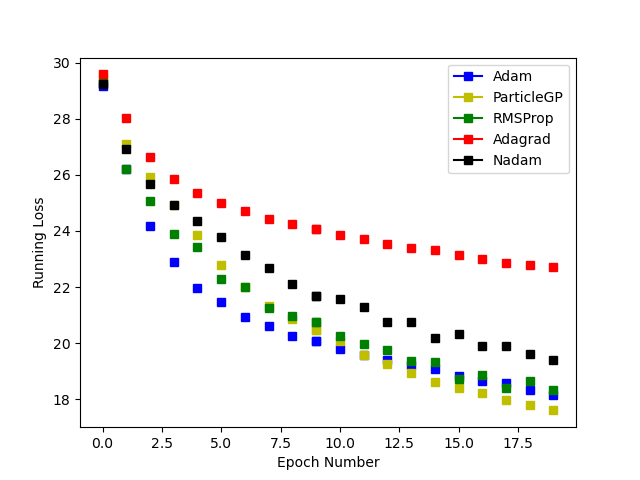}}
  \hfill
  \subfloat[Training Accuracy]{\includegraphics[width=0.5\textwidth]{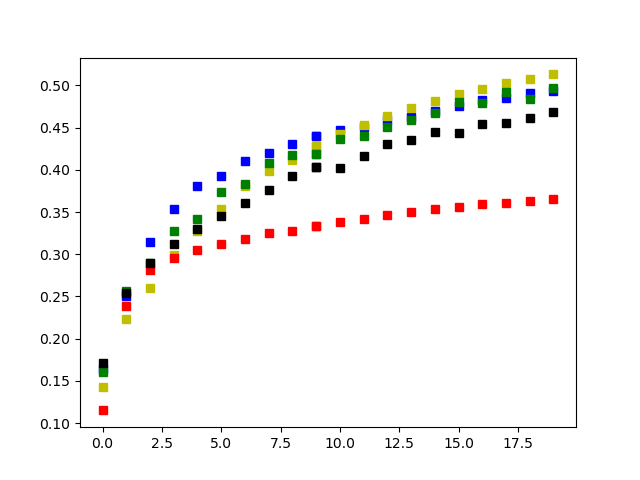}}
  \caption{Performance on the CIFAR10 dataset}
  \label{fig:CIFAR10}
\end{figure}

To justify that the ParticleGP algorithm can handle high dimensional parameter space, we test it using a simple neural network architecture involving CNNs in the Computer-Vision based classification task and compare it's performance to state-of-the-art optimizers in the field. The datasets explored are: CIFAR-10, MNIST, Fashion-MNIST. The number of epochs all optimizers were run was $20$, batch size of images used for training was $4096$, the learning rate for all optimizers was fixed to be $0.001$ and the loss function used was CrossEntropyLoss. For particleGP, we used same settings as above, except the number of particles ($N$) are $20$. Additionally, after gradient approximation, we applied the Adam updates to compute values of parameters for the next iteration. Table \ref{table:4} showcases the comparison. The figures \ref{fig:MNIST}, \ref{fig:FashionMNIST}, and \ref{fig:CIFAR10} illustrate plots of the running loss and training accuracy. We observe that ParticleGP had the highest or the second-highest accuracy and running loss out of all optimizers. The plots showcase that despite ParticleGP being slower in the beginning, it catches up significantly and outperforms others at the end.

\begin{table}
\caption{Comparison of the optimizers on classification datasets based test-accuracy and Cross-Entropy loss}
\centering
\begin{tabular}{||c |c | c | c | c | c | c||} 
 \hline
 Dataset & Metric & ParticleGP & Adam & RMSProp & NAdam & AdaGrad\\

 \hline
 \multirow{2}{*}{MNIST} & Testing Accuracy & \textbf{92.43\%} & 91.73\%  & 89.98 \% & 91.37\% & 79.11\% \\ 
   \cline{2-7}
  & Loss & \textbf{0.7694} & 0.8280 & 0.9766 & 0.8751 & 2.9356\\
\hline
 \multirow{2}{*}{Fashion MNIST} & Testing Accuracy & 83.80\% & 83.78\%  & \textbf{83.85\%} & 82.31\% & 74.12\% \\ 
   \cline{2-7}
 & Loss & 1.3626 & 1.3558  & \textbf{1.3326} & 1.4435 & 2.5767\\
 \hline
 \multirow{2}{*}{CIFAR-10} &  Testing Accuracy & 50.01\% & 49.17\%  & \textbf{50.47\%} & 45.74\% & 36.39\%\\ 
  \cline{2-7}
 & Loss & \textbf{4.1296} & 4.2224 & 4.2047 & 4.5503 & 5.2040\\
 \hline
\end{tabular}
\label{table:4}
\end{table}

Based on these results, we can infer that ParticleGP performs as good as or sometimes slightly better than the current state-of-the-art optimizers in classification tasks.

\newpage

\section{Conclusion}
\label{sec:limitations}

In this research, we proposed the Particle-based Gaussian process Optimizer, a novel technique utilizing Gaussian process (GP) regression, called as ParticleGP, for representing the dynamical process of gradient descent. In particular, the GP regressed over computed gradients at the neighbourhood points to predict the best direction of descent for the network parameters in every iteration of the scheme. This approach, motivated from utilizing gradients in standard optimizers and using evaluations from a swarm of particles from particle swarm optimization (pso) methods, attempts to incorporate the best of both techniques. The usage of gradients proved essential to ensure decent along the fastest direction, while the swarm of particles increased the exploration for detecting and converging to the global minima. We described the details of the multi-output Gaussian process regression for predicting gradients and summerized our approach in a single algorithm. After testing ParticleGP on $10$ distinct non-convex quadratic optimization problems, we observed it had a slightly better performance than other optimizers. We also tested ParticleGP on a computer vision task and found equivalent to slightly better results than conventional optimizers. The biggest limitation of this work is scalability since introducing Gaussian process regression in the optimizer considerably increases required computational resources and makes it challenging to used in heavier neural network architectures. 

\paragraph{Limitations and Future Work}

\begin{enumerate}
\item ParticleGP, since it uses a Gaussian process Regression model for predicting gradients after each iteration, it requires heavy computational resources for preparing the model and computing the matrix inverses for the posterior distribution update; and whether it showcases a performance tantamount to the same remains to be seen. In the appendix, we provide an approach to utilize sparse-Gaussian process
\item The hyper-parameters \textit{learning\_rate}, \textit{swarm\_size}, variance for initializing particles directly impact the approximation of gradients for the main agent, and thus are extremely sensitive. A slight change affects the overall results by a significant margin. Hence, a mechanism to optimally control these hyper-parameters; which we are currently working on, is essential. The first sub-section in appendix provides motivation for a control-based gradient-descent technique.
\item The experiments used a relatively light model for computations. It is unclear how ParticleGP would perform in domain-specific models for involving heavy data-sets like VGG, Resnet, Transformer, ViT, etc.
\end{enumerate}

\bibliographystyle{ieee_fullname}
\bibliography{references}

\end{document}